\documentclass[sigconf]{acmart}
\AtBeginDocument{%
  }

\acmConference[]{\ }{\ }{\ }
\renewcommand\footnotetextcopyrightpermission[1]{} 
\usepackage{array}
\usepackage{booktabs}
\usepackage{caption}
\usepackage{subcaption}

\begin{document}

\title{Scaffolding Creativity: Integrating Generative AI Tools and Real-world 
Experiences in Business Education}


\author{Nicole C. Wang}
\email{cw3715@nyu.edu}
\affiliation{%
  \institution{New York University Shanghai}
  \city{Shanghai}
  \country{China}
}

\renewcommand{\shortauthors}{Wang}

\begin{abstract}
  This exploratory study investigates the intersection of Generative AI tools and experiential learning in business education. Through a case study of an innovative undergraduate course, we examine how students interact with and adapt to various AI modalities—from text-based tools to image generation—alongside real-world experiences. Our findings reveal how this integrated approach enables novice users to overcome creative barriers, accelerates skill acquisition, and creates a dynamic interplay between AI-generated insights and real-world validation. We identify critical interaction challenges, including prompt engineering patterns and the need for more intuitive AI interfaces in educational contexts. These insights inform the design of future AI tools for creative learning and contribute to broader HCI discussions about human-AI collaboration in educational settings.
\end{abstract}


\thanks{This is the author’s version of the work accepted for publication in \textit{CHI EA ’25: Extended Abstracts of the CHI Conference on Human Factors in Computing Systems}. The final published version is available at \url{https://doi.org/10.1145/3706599.3720283}}


\maketitle

\section{Introduction}
The rapid evolution of Generative AI tools is fundamentally transforming how we approach teaching and learning across disciplines. While current research has explored various applications of AI in education, including personalized learning systems and automated assessment tools\cite{Limo2023,Fuchs2023,Mizumoto2023}, little attention has been paid to how these tools can scaffold complex, interdisciplinary learning experiences. This gap is particularly evident in business education, where students must increasingly navigate the intersection of creativity, technology, and business strategy.

The challenge is multifaceted: students need to master not only business fundamentals but also develop the creative and technical capabilities demanded by an AI-augmented business environment. Traditional business education, with its discipline-by-discipline approach\cite{Currie2003}, creates a disconnect between siloed learning and the integrative nature of real-world practice. This disconnect is further amplified by vast variations in students' AI literacy\cite{Hornberger2023} and their ability to effectively leverage rapidly evolving AI tools in creative and strategic tasks.

Novice users also face significant hurdles in prompt engineering \cite{ZamfirescuPereira2023, Sedlbauer2024}. The rapid deployment of advanced language models like GPT-4 and Claude 3.5 Sonnet, alongside image and video generation models such as Midjourney and Runway, presents educators with a moving target. While students often have access to these tools, their proficiency in leveraging them for creative and entrepreneurial tasks varies widely. Moreover, most educational research has focused primarily on text-based models, leaving a significant gap in understanding how to effectively integrate multimodal AI tools into experiential learning.

To address these challenges, this case study serves as exploratory research to understand how the integration of AI tools with real-world experiences impacts students' creative processes and learning outcomes in business education. While the study does not employ formal pre- and post-assessments or control groups due to the nature of the course setting, and the sample is limited, this initial investigation aims to identify promising directions for future research in this rapidly evolving space: 

\begin{enumerate}
    \item How students leverage different AI tools to overcome creative barriers 
    \item The interplay between AI-generated insights and real-world observations
    \item The role of AI tools in scaffolding the development of visual and creative skills
\end{enumerate}

Creativity, broadly defined as the ability to generate ideas that are both novel and effective \cite{Runco2012}, plays a central role in this study, as AI tools have the potential to redefine how students engage with idea generation, problem-solving, and execution. Business creativity is often market-oriented, manifesting through strategies, business models, and market solutions \cite{Berglund2006}. This historically makes it more challenging to evaluate creativity’s effectiveness in real-world contexts. In this study, we redefine ‘effectiveness’ in creativity not only as practical feasibility but also as the ability to bridge conceptual thinking with real-world execution, with the aid of AI tools.

Our findings reveal that this integrated approach not only accelerates knowledge acquisition but also enables students to overcome traditional creative barriers, create more tangible non-text-based deliverables, and develop more nuanced understanding of their business ideas through the dynamic interaction between AI tools and real-world experiences. We also identify key challenges, including the need for new assessment methods and the rapid evolution of AI capabilities creating a moving target for curriculum design.

These insights contribute to ongoing HCI discussions about human-AI collaboration in learning contexts and provide actionable recommendations for designing educational experiences that effectively combine AI tools with real-world practice. As an exploratory study, our work not only documents initial findings in this emerging space but also aims to catalyze future research directions in the design of AI interfaces that better support creative learning and interdisciplinary skill development. The patterns and challenges identified here, while preliminary, point to rich opportunities for more extensive investigation by the HCI community.

\vspace{1em}
\section{STUDY DESIGN AND METHODOLOGY}

\subsection{Course Overview}

The course \emph{Reinventing the Brand} evolved from initially focusing on reimagining existing brands to having students create their own brands, recognizing the increasing prevalence of individual-created brands and the potential of AI to lower barriers to entry. This shift aimed to prepare students for a future where AI might enable small business owners to compete more effectively in branding and content creation \cite{West2018,Wang2023}.

Students were required to produce tangible outputs including:brand websites with logos and designs, social media posts and content strategies, brand development presentations, AI-generated commercial videos and physical artifacts (eg. packaging prototypes, fabric samples).

The curriculum adopted a \emph{breadth-over-depth}  approach to provide a comprehensive overview of brand creation and management in the AI era. This design deliberately broke down traditional business education silos, allowing for personalized interdisciplinary learning trajectories that reflect the integrative nature of modern brand development.

\begin{table}[b]
  \caption{Participants}
  \label{tab:student-info}

    \begin{tabular}{lll}
      \toprule
      ID & Year & Major \\
      \midrule
      S1 & Senior & Traditional Business \\
      S2 & Sophomore & Interdisciplinary (Business focus) \\
      S3 & Junior & Interdisciplinary (Business focus) \\
      S4 & Sophomore & Interdisciplinary (Design focus) \\
      S5 & Sophomore & Interdisciplinary (Design focus) \\
      S6 & Freshman & Undeclared \\
      S7 & Junior & Interdisciplinary (Business focus) \\
      S8 & Freshman & Undeclared \\
      \bottomrule
    \end{tabular}

  \vspace{2mm} 
  \parbox{70mm}{ 
      \footnotesize *S7 and S8 completed the final project as a group.
  }

\end{table}

\subsection{Participants}
The course enrolled a total of 8 students (Table 1), representing a diverse international cohort from Europe, North America, and various regions of Asia, encompassing six different nationalities. 

Despite the presence of some students with business-related coursework, the overall background in business among the participants was limited. Importantly, while all students reported that they had never received any formal AI literacy training prior to this course, they exhibited varying levels of engagement experience with ChatGPT-3.5.

\vspace{1em}
\subsection{Phased Integration of AI Tools and Real-World Experiences}

\newcommand{\customphase}[1]{%
  \vspace{2em}
  {\large\bfseries #1}
  \vspace{0.3em}
}

\vspace{-0.8em}
\customphase{Phase 1: Text-based AI (ChatGPT-3.5)}

\noindent Students were encouraged to use ChatGPT throughout the course. At the beginning of the semester, most students used ChatGPT 3.5, as GPT-4o was not freely accessible until mid-semester (capped at 5 free messages per day for the free tier). The use of ChatGPT was optional, leading to varied adoption rates among the students.

\vspace{1em}
\noindent\customphase{{Phase 2: Real-World Experience Integration}}
\vspace{-2em} 

\noindent The course integrated two key real-world experiences to complement AI-assisted learning. Initially, students visited a local shopping mall to observe brand strategies and consumer behaviors, applying foundational business concepts to retail environments. This consumer-facing experience set the stage for understanding the broader supply chain.

 The centerpiece of experiential learning was a two-day field trip to Yiwu Small Merchandise Market in Zhejiang Province, China's hub for small commodity wholesale market, which spans 5.5 million square meters with over 75,000 shops, attracting 500,000+ international buyers yearly. This immersion exposed students to global trade dynamics and emerging consumption trends.

 This experiential learning was later complemented by students re-engaging with manufacturers via online wholesale platforms. With their newly acquired knowledge and insights, students leveraged these interactions to finalize their brand development projects.

\vspace{0.6em}
\noindent This real-world component served a \emph{dual purpose}: 

\begin{itemize}
  \item \textbf{Opportunity to validate AI-generated insights:} Students could test the assumptions and insights generated by AI during the local trip, creating a dynamic interplay between AI-generated output and real-world experiences. 

  \item \textbf{Experiential learning:} Students experienced firsthand the nuances of human interaction in business settings and the value of experiential knowledge – aspects that AI tools currently cannot fully replicate.
    
\end{itemize}

\vspace{0.6em}
\noindent\customphase{{Phase 3: Multimodal AI Introduction}}
\vspace{-2em} 

\noindent To facilitate the creation of tangible outputs such as brand images and AI video commercials, we introduced a Basic GenAI training session in Week 9. This session covered coomercial AI tools such as Midjourney, Leonardo.AI, Pika, Suno.ai, ElevenLabs. 

\newcolumntype{C}[1]{>{\centering\arraybackslash}p{#1}}

\begin{table*}
\caption{Case Study Focus}
\label{tab:project-components}

\begin{tabular}{p{4cm}p{7cm}C{2cm}}
\toprule
Scope of Focus & Details & Sample size (N) \\
\midrule
Mini-Presentation 4 (MP4) & Research on raw materials and production & 8 \\
MP4 ChatGPT documentation & Prompt history from three students & 3 \\
Field Trip to Yiwu & Observation and reflection session & 8 \\
Branding Proposal & Initial proposal for the brand concepts (text-based)\footnote{a} & 7 \\
Midjourney documentation & Interaction history from 4 Students & 4 \\
Final Brand Portfolio & Holistic analysis of the AI-generated elements & 8 \\
\bottomrule
\end{tabular}

  \vspace{2mm} 
  \parbox{140mm}{ 
      \footnotesize\textsuperscript{1} Students were encouraged to submit their prompt histories but are not required, resulting in an even smaller sample size for ChatGPT and Midjourney prompt analyses.
  }

\end{table*}

\subsection{Scope of Focus and Research Methodology}
The course's assignments include eight mini-presentations (MPs), a midterm branding proposal, and a brand portfolio, which includes a presentation, brand website, social media posts as well as an AI video commercial. The MPs serve the purpose of scaffolding students' knowledge and skills to create their final culminating project.

Given the exploratory nature of this study, we documented student interactions with AI tools across the following components (Table 2):

\begin{enumerate}
    \item \textbf{AI-assisted learning in unfamiliar domains:} We examined MP4 outputs and available prompt histories ($N=3$) to understand how students used ChatGPT to research unfamiliar topics.
    
    \item \textbf{Integration of AI insights with real-world experiences:} We compared students' AI-generated knowledge before the Yiwu trip with  their post-trip in-class reflections, based on observational notes and discussions with students before, during, and after the visit.
    
    \item \textbf{AI-assisted visual transformation:} We analyzed the evolution of students' brand concepts from text-based proposals to visual portfolios. This included examining Midjourney interaction history ($N=4$) and comparing initial brand proposals with final visual presentations.
\end{enumerate}

\vspace{\baselineskip}
{\noindent\large\bfseries Data Analysis Approach and Limitations}

\noindent Due to the constraints of conducting research within an active course setting, formal experimental designs such as pre/post assessments or control groups were not feasible. Additionally, given the small sample size and the exploratory nature of this study, we did not employ structured qualitative analysis methods such as thematic analysis or inductive coding. Instead, our findings were drawn from observed patterns, student-reported experiences, and qualitative comparisons of AI-assisted outputs and non-AI-assisted outputs in previous similar courses. Specifically:

\begin{itemize}
  \item \textbf{Prompt Histories} were analyzed through instructor observations of how students refined their prompts and iterated on AI-generated content. Given the small sample size, this analysis remains an interpretative assessment rather than a formally coded dataset. Some students explicitly described their decision-making process in reflections, which provided additional qualitative insights.

  \item \textbf{Project Output} were reviewed to identify differences in creativity and execution. While this study does not conduct a formal comparative analysis, the instructor qualitatively compared branding portfolios from the current AI-assisted course with branding portfolios from a prior similar course that did not use AI. However, these insights are based on instructor experience rather than controlled experimental comparisons, and future studies could incorporate structured comparative methods to assess AI’s impact on creative capacity more systematically.

  \item \textbf{Classroom Observations \& Discussions} were informally documented through instructor interactions with students during the course. However, these insights were not systematically logged, as they emerged through ongoing discussions rather than structured data collection. Future research will aim to capture these observations more systematically through structured field notes or follow-up interviews.

\end{itemize}

While this study does not formally measure whether AI enhanced students’ creative capacity or merely improved execution efficiency, instructor observations suggest that students engaged AI for both ideation and execution. However, the extent and impact of these effects remain unclear and require further research.

\section{FINDINGS}

\subsection{AI-Assisted Research in Unfamiliar Topics}
In MP4, students investigated production processes for their brand concepts to understand how product features are affected by raw materials, production and supply chain dynamics, which are not typically taught in business programs. This assignment was strategically designed to observe how students utilize ChatGPT to research unfamiliar topics typically not covered in traditional business curricula.

\vspace{0.3cm}
\noindent The assignment instructions included general ChatGPT usage principles (Appendix B):

\begin{itemize}
    \item encouraged open-ended exploration
    \item start with broad questions to establish a foundation
    \item take an iterative approach
\end{itemize}

\noindent An additional aspect of the assignment instructions was the deliberate \emph{de-emphasis} on accuracy. Unlike typical academic requirements, students were explicitly told that ``100\% accuracy'' was not the goal. This approach aimed to encourage engagement with unfamiliar topics and facilitate broader exploration without the constraints of constant fact-checking when time is limited.

While all students used ChatGPT-3.5 for research, only three documented their prompt histories, providing some insights into different learning patterns:

\vspace{\baselineskip}
{\noindent\large\bfseries Non-Linear Learning Trajectories and Knowledge Integration}
\vspace{0.3em}

\noindent Despite the assignment suggesting a structured approach from broad to specific inquiries (Appendix B), students demonstrated distinctly different patterns of AI interaction based on their domain knowledge. These patterns reveal how students adaptively integrated AI assistance with their existing expertise and external resources.

Students with domain expertise demonstrated sophisticated, non-linear approaches to knowledge building. For instance, S1, leveraging garment production experience, began with specific material inquiries ("Which items of clothing can be made with 100\% hemp fabric?"[Appendix A, S1, P1]) rather than broad material exploration. This approach indicates an attempt to validate and expand existing knowledge using ChatGPT as a reference point. 

In contrast, S2 drawing on design experience, initiated the interaction by describing her product vision with clear design-oriented specifications such as “water bottle pockets” and “enough space for students”[S2,P1]. When this  prompt didn't yield the desired response, S2 adapted by returning to broader material questions.  

S3, who had limited domain experience, followed a more systematic approach aligned with the assignment suggestions. This pattern suggests that students without prior domain knowledge may rely more heavily on the structured approach provided in the instructions.

Students’ background in domain knowledge also seemingly affected their \textit{analytical approach} when interacting with ChatGPT. Students with less domain knowledge tend to accept AI responses more readily, using them as scaffolding to build general understanding. But the students with certain domain knowledge are more likely to demonstrate a more critical and nuanced engagement with AI-generated information. For example, when S1 learned that hemp can be used to create “even jeans” she immediately followed up with a question [S1,P4] about the differences between cotton and hemp jeans for deeper inquiry. Similarly, S2 cited information from internet sources[S2,P4], particularly when ChatGPT's answers seemed too theoretical for her practical, production-oriented questions.

Finally, S1 demonstrated a tendency to inquire about \textit{tangential} knowledge. In final queries, S1 focused on ramie [S1, P7-P8], a fabric not mentioned in previous responses. This sudden shift in focus could represent either a recollection of prior knowledge or genuine curiosity sparked by the AI interaction. This behavior could suggest the AI's responses might trigger memories of related concepts or introduce new ideas that pique the student's interest, leading to unexpected learning paths.

The diverse learning trajectories observed in these cases illustrate how AI tools can facilitate non-linear, adaptive learning processes by enabling students to dynamically integrate new information with prior knowledge and external resources \cite{Rasul2023,Rudolph2023}. These interactions reveal the potential of AI to scaffold both novice and expert learners. However, the same flexibility that supports adaptive learning also introduces the risk of scattered focus if learners lack guidance. This highlights the importance of designing AI systems that not only provide information but also guide users in structuring their inquiry and critically evaluating responses. 

\vspace{\baselineskip}
{\noindent\large\bfseries \textbf{\textbf{Accelerated Domain-Specific Knowledge Acquisition}
}
}
\vspace{0.3em}

\noindent Another interesting observation across all eight students' presentations was the rapid development of project-specific knowledge in product development and manufacturing. The prompt histories from the 3 samples revealed that students could navigate technically-oriented questions and answers and complete their research process within 4-8 prompts.

This observation demonstrates the potential of AI as a tool for just-in-time learning in project-based contexts \cite{Seo2021,Kim2024}, where students can quickly acquire specific expertise needed for their individual projects within compressed timelines. However, this accelerated knowledge acquisition raises important considerations about the balance between speed, depth and retention in learning \cite{Rudolph2023, Seo2021}.

\subsection{\textbf{Dynamic Interaction with Real-World Experiences} }
Our integration of AI tools with real-world experiences revealed interesting patterns in how students navigate between digital and physical learning environments, particularly in validating and adapting AI-generated knowledge.

\vspace{\baselineskip}
{\noindent\large\bfseries \textbf{\textbf{\textbf{Human-AI Collaboration in Complex Environments}
}
}
}
\vspace{0.3em}

\noindent After their AI-assisted research into raw materials and production processes, students successfully identified potential manufacturers on 1688.com, a major B2B e-commerce platform, demonstrating how AI can expand novice users' capabilities in complex market research tasks typically reserved for experienced professionals.

A significant insight emerged from how students combined AI-generated knowledge with real-world experiences in the Yiwu market. The rapid knowledge acquisition facilitated by AI enabled students to effectively engage in vendor interactions, with some successfully presenting themselves as experienced buyers. This was particularly notable as vendors typically identify and offer unfavorable deals to novice buyers who don't ask the "right" questions. The AI-assisted preparation seemed to have provided students with sufficient domain knowledge to ask pertinent questions and engage confidently in nuanced conversations.

The interplay between AI preparation and human adaptability suggests a novel approach in business experiential learning. Students used AI-generated knowledge as a foundation, which they dynamically adjusted based on real-time observations and interactions. This continuous recalibration between AI outputs and human experiences reflects ongoing discussions on Human-AI collaboration in learning contexts \cite{Nah2023}.

\vspace{\baselineskip}
{\noindent\large\bfseries \textbf{\textbf{\textbf{Uncovering AI limitations in Real-World Engagement}
}
}
}

\vspace{0.2em}
\noindent However, the real-world engagement also revealed limitations in AI-generated knowledge.  For example, S1's discovery that 100\% hemp production was far less common than AI had implied. Moreover, many students encountered a wide array of materials and production techniques during their visit that were not mentioned in their AI-assisted research. 

This discovery points to two known limitations of AI-generated knowledge: first, its tendency to provide generic answers unless prompted with very specific context \cite{Memarian2023}, which is difficult for student users; and second, its potential lag in incorporating the latest innovations or highly specialized industry knowledge \cite{Zhou2023}. This suggests a need for AI interfaces that can incorporate real-world validation mechanisms and support iterative refinement of knowledge based on user feedback.

\subsection{AI-Assisted Visual Transformation}
\noindent Traditionally, business school assignments have relied on text-based presentations supplemented with basic visuals, partly due to students' non-design backgrounds. Our study investigates how the introduction of multimodal AI tools transforms students' ability to visualize and communicate business concepts. Four students (N=4) maintained detailed records of their Midjourney interactions in the class Discord server after the training session, while others used various AI image generation tools through personal accounts. Table 3 presents the statistics of these Midjourney interactions, including initial concept prompts, iterative prompts (usually style or component changes), upscaling actions, remixes, uploads,  image-to-text descriptions and command error corrections.

\vspace{\baselineskip}
{\noindent\large\bfseries \textbf{AI as a Cognitive Scaffold and Human-AI Co-Creation}
}

\vspace{0.2em}
\noindent Our study revealed that AI tools functioned not merely as implementation aids, but as collaborative partners in both concept development and visualization. This partnership manifested as an iterative process where AI-generated visuals influenced conceptual thinking, which in turn guided further visual exploration. In particular, the introduction of multimodal AI tools after midterm appeared to serve as cognitive scaffolds which was particularly evident in the expansion and refinement of initial ideas:

\vspace{0.5em}
 \textit{Concept Expansion and Refinement:} Visual presentations often expanded upon initial text-based ideas, suggesting that the AI-assisted visual creation process prompted students to further refine and develop their brand concepts.

 \textit{Nuanced Expression:} The visual medium allowed students to express nuanced ideas that were difficult to convey through text alone, indicating that AI tools may be facilitating more complex conceptual thinking.
\vspace{0.7em}

\noindent For example, S3's brand concept aimed to "cross the line between femininity and masculinity," a nuanced idea difficult to convey through text alone. Figure 1 showcases the concept images S3 generated in an iterative cycle on Midjourney for her brand portfolio and AI commercial. In post-class correspondence, S3 detailed her iterative prompting process:

\vspace{\baselineskip}
\emph{ "… was trying to display feminine hands doing things that typically men are associated with the most: such as playing the drums and fixing a car”. } But because PikaLabs she used had a tendency to generate extra fingers, she \emph{“turned to generating images of feminine hands fixing a car.”}
\vspace{\baselineskip}

 This iterative process led S3 to have deeper reflections on her brand's message and target audience:

\vspace{\baselineskip}
\emph{ "I realized my initial concept might appeal to an unintended audience…I generated images of a man and a woman fixing cars together to try to promote what I believe is a better envisionment of gender harmony/equality”}
\vspace{\baselineskip}

The iterative visual exploration not only helped S3 refine her brand concept but also prompted a more nuanced refinement of her target audience definition and brand values. 

Similarly, S2's interaction with AI revealed how unexpected elements in generated images could spark new creative directions. After noticing a blurry figure in one AI-generated image, she incorporated human figures in subsequent prompts, better representing her brand's emphasis on ease of use.

 This collaborative process aligns with the vision that AI tools act not just as implementers of human ideas, but as active collaborators in the creative process \cite{Kim2022, Yin2023}. It also resonates with the notion of AI as a coach or a colleague \cite{Lubart2005}, stimulating new directions of thought and encouraging students to reconsider and enhance their initial concepts.

\begin{table}
  \caption{Midjourney Interactions (Post-Training)}
  \label{tab:midjourney-interactions}
  \centering
  \begin{tabular}{lrrrr}  
    \toprule
    \textbf{ID} & \textbf{S2} & \textbf{S3} & \textbf{S4} & \textbf{S5} \\
    \midrule
    Interactions & 90 & 31 & 40 & 12 \\
    Concepts & 14 & 10 & 12 & 4 \\
    Iterations & 45 & 10 & 19 & 6 \\
    Effective Interactions per Concept & 2.6 & 1.0 & 1.6 & 1.5 \\
    \bottomrule
  \end{tabular}

  \vspace{2mm} 
  \parbox{77mm}{ 
      \footnotesize\textsuperscript{1} Students could make errors in their prompts due to their unfamiliarity with the command format in Midjourney's server on Discord, causing ineffective interactions.
  }

\end{table}

\vspace{\baselineskip}
{\noindent\large\bfseries \textbf{Democratization of Visual Design Skills}
}
\vspace{0.3em}

\noindent The rapid adaptation to visual communication tools suggests that multimodal AI is democratizing design capabilities in non-design disciplines\cite{Adetayo2024,Yin2023}. All students, regardless of prior design experience, demonstrated strong capability in visual communication, producing professional-quality outputs that markedly exceeded typical business course submissions.

This democratization was particularly notable given that none of the students had prior experience with text-to-image or image-to-video tools in professional contexts before the Week 9 training session. The rapid skill acquisition suggests that current AI interfaces may be tapping into an innate capacity for visual expression previously constrained by technical barriers. This raises important implications for interface design. How might AI tools better support the transition from textual to visual thinking? What interface elements best facilitate creative exploration for non-designers? How can AI tools maintain professional output quality while supporting user agency and learning?

\begin{figure*}[!htb]
\caption{S3's Interactions with Midjourney Reflect Brand Concept Refinement}
\vspace{1em}  
\label{fig:s3-process}
\centering
\begin{subfigure}[t]{0.3\textwidth}
    \centering
    \includegraphics[width=\textwidth]{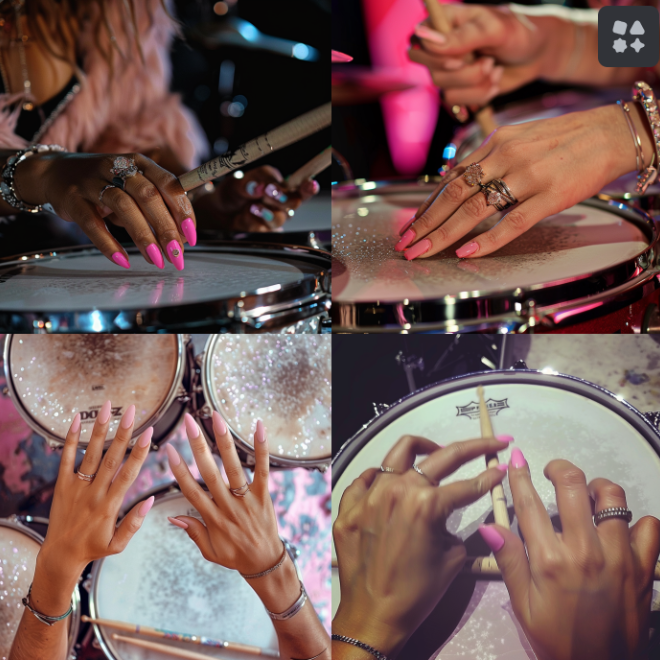}
    \caption*{1st concept: \textit{a first person POV realistic picture of a pair of hands with long pink nails playing the drums}}  
    \label{fig:drum}
\end{subfigure}
\hfill
\begin{subfigure}[t]{0.3\textwidth}
    \centering
    \includegraphics[width=\textwidth]{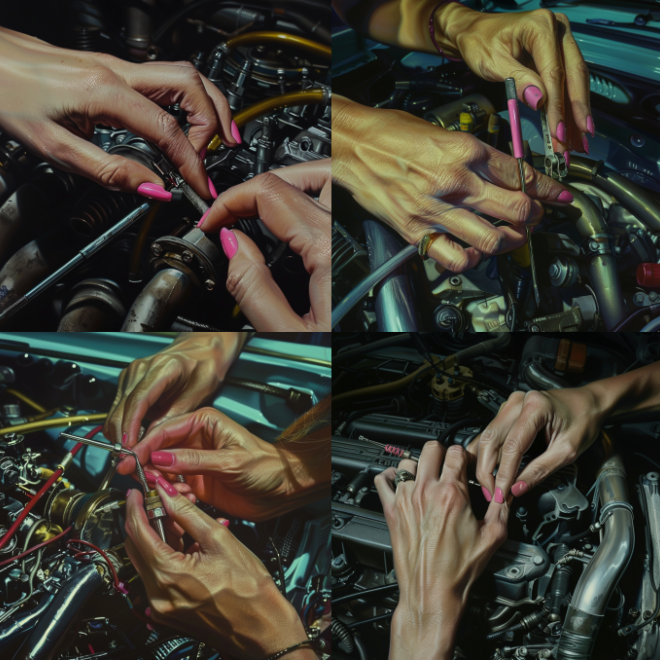}
    \caption*{2nd concept: \textit{a first person POV realistic picture of a pair of hands with long pink nails holding a screwdriver while fixing a car's engine}}  
    \label{fig:car}
\end{subfigure}
\hfill
\begin{subfigure}[t]{0.3\textwidth}
    \centering
    \includegraphics[width=\textwidth]{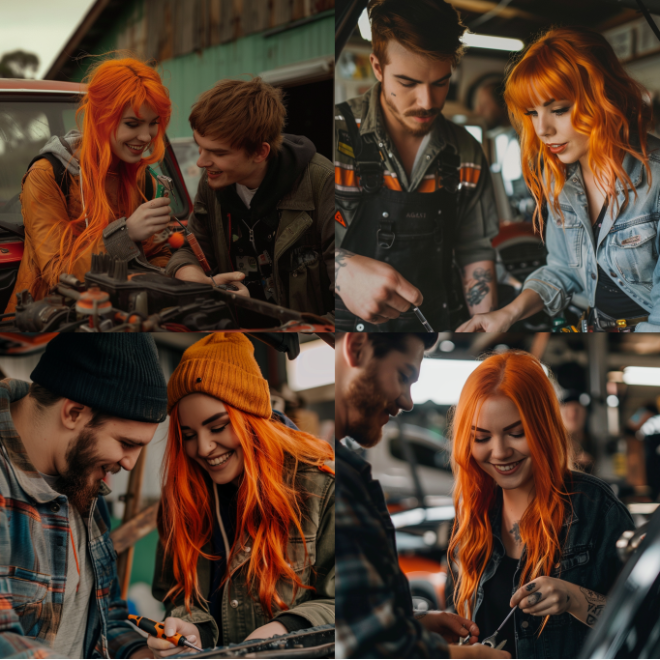}
    \caption*{3rd concept: \textit{a realistic photo shoot of a woman with orange hair holding a screwdriver while happily helping her guy friend fix his car}}  
    \label{fig:manwoman}
\end{subfigure}
\end{figure*}

\vspace{0.5cm} 
\noindent\customphase{{Shift Towards Individual Expression and Engagement}}
\vspace{-0.5cm} 

\noindent Despite encouragement to form pairs after the brand proposal stage, six out of eight students insisted on working individually. The ability to rapidly generate and refine designs independently may have reduced the perceived necessity of collaboration, as students no longer relied on peers to fill skill gaps. This aligns with the concept of AI as an Empowering Partner\cite{UNESCO2023}, in which AI tools act as individualized creative supports, fostering a sense of agency and competence. Additionally, this high motivation may resonate with Csikszentmihalyi’s Flow Theory\cite{Csikszentmihalyi1989}, suggesting that AI tools can potentially create an optimal challenge-skill balance that facilitates a state of flow where students felt both capable and motivated to pursue ambitious individual projects.

However, this shift toward individual learning raises important considerations for business education, where teamwork and interpersonal skills are crucial for professional success \cite{Rasul2023}. While AI enhances technical execution and creative iteration, it does not inherently foster the interpersonal abilities necessary for collaboration, negotiation, and shared decision-making—skills that traditionally develop through direct peer interaction. This suggests that AI-driven workflows may unintentionally deprioritize teamwork, potentially altering how students experience and develop collaborative competencies. Future course designs should explore how AI can be integrated without diminishing opportunities for interpersonal learning and teamwork.

\section{IMPLICATIONS AND LIMITATIONS}
The emergence of non-linear learning patterns in AI interactions suggests the need for more adaptive interfaces that can accommodate diverse learning approaches. Our observations reveal that users navigate AI capabilities in more complex ways, with their domain knowledge and creative goals significantly influencing their interaction patterns. This finding calls for new interface paradigms that can dynamically adjust to users' varying expertise levels and learning trajectories. Such adaptive interfaces are particularly relevant for the design of personalized (or self-directed) learning systems, which currently tend to focus primarily on AI model capabilities and knowledge integration rather than supporting the complex, non-linear nature of creative exploration and skill development. Future AI systems could proactively suggest real-world tasks or experiments tailored to a user's specific progress and interactions.

The interplay between AI-generated knowledge and real-world validation experiences suggests an important direction for future interface development. Current AI systems largely operate as closed loops, with limited ability to incorporate real-world feedback. Our findings indicate a need for interfaces that can not only integrate observational data from real-world experiences but also adapt iteratively based on this feedback. This iterative refinement is especially critical in educational contexts where theoretical knowledge must align with practical realities. Moreover, the adaptive and iterative interaction patterns observed in this study could inform the design of AI tools for professional and lifelong learning, where learners must continuously apply knowledge in evolving real-world scenarios.

From a broader perspective, our findings can be situated within Wu et al.'s Human-AI Co-Creation Model\cite{Wu2021}. It provides a useful conceptual foundation for understanding how humans and AI interact through cycles of perception, thinking, expression, collaboration, building, and testing. For instance, AI facilitates deeper and wider exploration during the thinking and expression phases, offering adaptive learning that supports ideation and concept development. This is then followed by real-world validation, where students build and test their knowledge by integrating AI-generated insights into practical observations and applications. Currently, AI’s role in perception remains limited, as it lacks the sensorimotor capabilities to directly interpret real-world contexts. Furthermore, while the authors of the model argue that humans and AI complement each other in the co-creative process, their relative strengths may shift across different learning contexts and evolve over time, Recognizing these complexities can inform the design of future AI learning systems that better support adaptive, iterative, and co-creative learning experiences.

Another promising area for future research emerges from our observations of individual creative work. The unexpected preference for individual work, despite the project's complexity, suggests that AI tools may be reshaping the role of collaboration in learning environments. While AI-driven workflows offer significant advantages, they may also unintentionally limit opportunities for students to develop interpersonal skills essential for professional collaboration.Future research should explore hybrid approaches to ensure that students benefit from both personalized AI assistance and human-centered collaborative learning experiences.

The methodological limitations of this study also point to opportunities for further investigation. Due to the constraints of conducting research within an active course setting, this study is exploratory in nature and does not employ formal qualitative coding techniques or controlled comparative analysis. Future research should incorporate structured analysis and even longitudinal assessment framework to examine AI’s impact on conceptual learning, critical thinking, and business knowledge development more systematically.

Given the small sample size and single-course context, further research should also examine how human-AI creative collaboration unfolds across different educational settings and disciplines. Additionally, the observed challenges with prompt engineering and AI bias—particularly in image generation—underscore the need for more sophisticated interface mechanisms that help users better articulate their creative intentions and critically evaluate AI outputs.

Finally, the study did not assess how mismatches between AI-generated outputs and students’ expectations impacted their trust in AI-generated knowledge. Future research should investigate how students develop AI literacy skills, including their ability to assess AI reliability, recognize biases, and refine AI-generated content critically \cite{Lee2024,Kim2024}.

These findings contribute to an emerging research agenda for the HCI community focused on developing more nuanced and effective approaches to human-AI collaboration in creative learning contexts. As educational institutions continue to embrace the integration of rapidly evolving AI tools, the development of interfaces that can effectively support these complex learning processes becomes increasingly critical.

\bibliographystyle{ACM-Reference-Format}
\bibliography{bib}


\begin{thebibliography}{26}


\ifx \showCODEN    \undefined \def \showCODEN     #1{\unskip}     \fi
\ifx \showDOI      \undefined \def \showDOI       #1{#1}\fi
\ifx \showISBNx    \undefined \def \showISBNx     #1{\unskip}     \fi
\ifx \showISBNxiii \undefined \def \showISBNxiii  #1{\unskip}     \fi
\ifx \showISSN     \undefined \def \showISSN      #1{\unskip}     \fi
\ifx \showLCCN     \undefined \def \showLCCN      #1{\unskip}     \fi
\ifx \shownote     \undefined \def \shownote      #1{#1}          \fi
\ifx \showarticletitle \undefined \def \showarticletitle #1{#1}   \fi
\ifx \showURL      \undefined \def \showURL       {\relax}        \fi
\providecommand\bibfield[2]{#2}
\providecommand\bibinfo[2]{#2}
\providecommand\natexlab[1]{#1}
\providecommand\showeprint[2][]{arXiv:#2}

\bibitem[Adetayo(2024)]%
        {Adetayo2024}
\bibfield{author}{\bibinfo{person}{Adebowale~Jeremy Adetayo}.} \bibinfo{year}{2024}\natexlab{}.
\newblock \showarticletitle{Reimagining learning through AI art: The promise of DALL-E and MidJourney for education and libraries}.
\newblock \bibinfo{journal}{\emph{Library Hi Tech News}}  \bibinfo{volume}{(ahead-of-print)} (\bibinfo{year}{2024}).
\newblock


\bibitem[Berglund and Wennberg(2006)]%
        {Berglund2006}
\bibfield{author}{\bibinfo{person}{Henrik Berglund} {and} \bibinfo{person}{Karl Wennberg}.} \bibinfo{year}{2006}\natexlab{}.
\newblock \showarticletitle{Creativity among entrepreneurship students: Comparing engineering and business education}.
\newblock \bibinfo{journal}{\emph{International Journal of Continuing Engineering Education and Life Long Learning}} \bibinfo{volume}{16}, \bibinfo{number}{5} (\bibinfo{year}{2006}), \bibinfo{pages}{366--379}.
\newblock


\bibitem[Csikszentmihalyi and LeFevre(1989)]%
        {Csikszentmihalyi1989}
\bibfield{author}{\bibinfo{person}{Mihaly Csikszentmihalyi} {and} \bibinfo{person}{Judith LeFevre}.} \bibinfo{year}{1989}\natexlab{}.
\newblock \showarticletitle{Optimal experience in work and leisure}.
\newblock \bibinfo{journal}{\emph{Journal of personality and social psychology}} \bibinfo{volume}{56}, \bibinfo{number}{5} (\bibinfo{year}{1989}), \bibinfo{pages}{815}.
\newblock


\bibitem[Currie and Knights(2003)]%
        {Currie2003}
\bibfield{author}{\bibinfo{person}{Graeme Currie} {and} \bibinfo{person}{David Knights}.} \bibinfo{year}{2003}\natexlab{}.
\newblock \showarticletitle{Reflecting on a critical pedagogy in MBA education}.
\newblock \bibinfo{journal}{\emph{Management Learning}} \bibinfo{volume}{34}, \bibinfo{number}{1} (\bibinfo{year}{2003}), \bibinfo{pages}{27--49}.
\newblock


\bibitem[Fuchs(2023)]%
        {Fuchs2023}
\bibfield{author}{\bibinfo{person}{Kevin Fuchs}.} \bibinfo{year}{2023}\natexlab{}.
\newblock \showarticletitle{Exploring the opportunities and challenges of NLP models in higher education: is Chat GPT a blessing or a curse?}. In \bibinfo{booktitle}{\emph{Frontiers in Education}}, Vol.~\bibinfo{volume}{8}. \bibinfo{publisher}{Frontiers Media SA}, \bibinfo{pages}{1166682}.
\newblock


\bibitem[Hornberger et~al\mbox{.}(2023)]%
        {Hornberger2023}
\bibfield{author}{\bibinfo{person}{Matthias Hornberger}, \bibinfo{person}{Anna Bewersdorff}, {and} \bibinfo{person}{Claudia Nerdel}.} \bibinfo{year}{2023}\natexlab{}.
\newblock \showarticletitle{What do university students know about Artificial Intelligence? Development and validation of an AI literacy test}.
\newblock \bibinfo{journal}{\emph{Computers and Education: Artificial Intelligence}}  \bibinfo{volume}{5} (\bibinfo{year}{2023}), \bibinfo{pages}{100165}.
\newblock


\bibitem[Kim et~al\mbox{.}(2022)]%
        {Kim2022}
\bibfield{author}{\bibinfo{person}{Jinhee Kim}, \bibinfo{person}{Hyunkyung Lee}, {and} \bibinfo{person}{Young~Hoan Cho}.} \bibinfo{year}{2022}\natexlab{}.
\newblock \showarticletitle{Learning design to support student-AI collaboration: Perspectives of leading teachers for AI in education}.
\newblock \bibinfo{journal}{\emph{Education and Information Technologies}} \bibinfo{volume}{27}, \bibinfo{number}{5} (\bibinfo{year}{2022}), \bibinfo{pages}{6069--6104}.
\newblock


\bibitem[Kim and Adlof(2024)]%
        {Kim2024}
\bibfield{author}{\bibinfo{person}{Minkyoung Kim} {and} \bibinfo{person}{Lauren Adlof}.} \bibinfo{year}{2024}\natexlab{}.
\newblock \showarticletitle{Adapting to the future: ChatGPT as a means for supporting constructivist learning environments}.
\newblock \bibinfo{journal}{\emph{TechTrends}} \bibinfo{volume}{68}, \bibinfo{number}{1} (\bibinfo{year}{2024}), \bibinfo{pages}{37--46}.
\newblock


\bibitem[Lee(2024)]%
        {Lee2024}
\bibfield{author}{\bibinfo{person}{Hyunsu Lee}.} \bibinfo{year}{2024}\natexlab{}.
\newblock \showarticletitle{The rise of ChatGPT: Exploring its potential in medical education}.
\newblock \bibinfo{journal}{\emph{Anatomical Sciences Education}} \bibinfo{volume}{17}, \bibinfo{number}{5} (\bibinfo{year}{2024}), \bibinfo{pages}{926--931}.
\newblock


\bibitem[Limo et~al\mbox{.}(2023)]%
        {Limo2023}
\bibfield{author}{\bibinfo{person}{Fernando Antonio~Flores Limo}, \bibinfo{person}{David Raul~Hurtado Tiza}, \bibinfo{person}{Maribel~Mamani Roque}, \bibinfo{person}{Edward~Espinoza Herrera}, \bibinfo{person}{José Patricio~Muñoz Murillo}, \bibinfo{person}{Jorge~Jinchuña Huallpa}, \bibinfo{person}{Victor~Andre Ariza~Flores}, \bibinfo{person}{Alejandro Guadalupe~Rincón Castillo}, \bibinfo{person}{Percy Fritz~Puga Peña}, \bibinfo{person}{Christian Paolo~Martel Carranza}, {and} \bibinfo{person}{José Luis~Arias Gonzáles}.} \bibinfo{year}{2023}\natexlab{}.
\newblock \showarticletitle{Personalized tutoring: ChatGPT as a virtual tutor for personalized learning experiences}.
\newblock \bibinfo{journal}{\emph{Przestrzeń Społeczna (Social Space)}} \bibinfo{volume}{23}, \bibinfo{number}{1} (\bibinfo{year}{2023}), \bibinfo{pages}{293--312}.
\newblock


\bibitem[Lubart(2005)]%
        {Lubart2005}
\bibfield{author}{\bibinfo{person}{Todd Lubart}.} \bibinfo{year}{2005}\natexlab{}.
\newblock \showarticletitle{How can computers be partners in the creative process: classification and commentary on the special issue}.
\newblock \bibinfo{journal}{\emph{International journal of human-computer studies}} \bibinfo{volume}{63}, \bibinfo{number}{4-5} (\bibinfo{year}{2005}), \bibinfo{pages}{365--369}.
\newblock


\bibitem[Memarian and Doleck(2023)]%
        {Memarian2023}
\bibfield{author}{\bibinfo{person}{Bahar Memarian} {and} \bibinfo{person}{Tenzin Doleck}.} \bibinfo{year}{2023}\natexlab{}.
\newblock \showarticletitle{ChatGPT in education: Methods, potentials, and limitations}.
\newblock \bibinfo{journal}{\emph{Computers in Human Behavior: Artificial Humans}} \bibinfo{volume}{1}, \bibinfo{number}{2} (\bibinfo{year}{2023}), \bibinfo{pages}{100022}.
\newblock


\bibitem[Mizumoto and Eguchi(2023)]%
        {Mizumoto2023}
\bibfield{author}{\bibinfo{person}{Atsushi Mizumoto} {and} \bibinfo{person}{Masaki Eguchi}.} \bibinfo{year}{2023}\natexlab{}.
\newblock \showarticletitle{Exploring the potential of using an AI language model for automated essay scoring}.
\newblock \bibinfo{journal}{\emph{Research Methods in Applied Linguistics}} \bibinfo{volume}{2}, \bibinfo{number}{2} (\bibinfo{year}{2023}), \bibinfo{pages}{100050}.
\newblock


\bibitem[Nah et~al\mbox{.}(2023)]%
        {Nah2023}
\bibfield{author}{\bibinfo{person}{Fiona Fui-Hoon Nah}, \bibinfo{person}{Ruilin Zheng}, \bibinfo{person}{Jingyuan Cai}, \bibinfo{person}{Keng Siau}, {and} \bibinfo{person}{Langtao Chen}.} \bibinfo{year}{2023}\natexlab{}.
\newblock \showarticletitle{Generative AI and ChatGPT: Applications, challenges, and AI-human collaboration}.
\newblock \bibinfo{journal}{\emph{Journal of Information Technology Case and Application Research}} \bibinfo{volume}{25}, \bibinfo{number}{3} (\bibinfo{year}{2023}), \bibinfo{pages}{277--304}.
\newblock


\bibitem[Rasul et~al\mbox{.}(2023)]%
        {Rasul2023}
\bibfield{author}{\bibinfo{person}{Tareq Rasul}, \bibinfo{person}{Sumesh Nair}, \bibinfo{person}{Diane Kalendra}, \bibinfo{person}{Mulyadi Robin}, \bibinfo{person}{Fernando de Oliveira~Santini}, \bibinfo{person}{Wagner~Junior Ladeira}, \bibinfo{person}{Mingwei Sun}, \bibinfo{person}{Ingrid Day}, \bibinfo{person}{Raouf~Ahmad Rather}, {and} \bibinfo{person}{Liz Heathcote}.} \bibinfo{year}{2023}\natexlab{}.
\newblock \showarticletitle{The role of ChatGPT in higher education: Benefits, challenges, and future research directions}.
\newblock \bibinfo{journal}{\emph{Journal of Applied Learning and Teaching}} \bibinfo{volume}{6}, \bibinfo{number}{1} (\bibinfo{year}{2023}), \bibinfo{pages}{41--56}.
\newblock


\bibitem[Rudolph et~al\mbox{.}(2023)]%
        {Rudolph2023}
\bibfield{author}{\bibinfo{person}{Jürgen Rudolph}, \bibinfo{person}{Samson Tan}, {and} \bibinfo{person}{Shannon Tan}.} \bibinfo{year}{2023}\natexlab{}.
\newblock \showarticletitle{ChatGPT: Bullshit spewer or the end of traditional assessments in higher education?}
\newblock \bibinfo{journal}{\emph{Journal of Applied Learning and Teaching}} \bibinfo{volume}{6}, \bibinfo{number}{1} (\bibinfo{year}{2023}), \bibinfo{pages}{342--363}.
\newblock


\bibitem[Runco and Jaeger(2012)]%
        {Runco2012}
\bibfield{author}{\bibinfo{person}{Mark~A. Runco} {and} \bibinfo{person}{Garrett~J. Jaeger}.} \bibinfo{year}{2012}\natexlab{}.
\newblock \showarticletitle{The standard definition of creativity}.
\newblock \bibinfo{journal}{\emph{Creativity Research Journal}} \bibinfo{volume}{24}, \bibinfo{number}{1} (\bibinfo{year}{2012}), \bibinfo{pages}{92--96}.
\newblock


\bibitem[Seo et~al\mbox{.}(2021)]%
        {Seo2021}
\bibfield{author}{\bibinfo{person}{Kyoungwon Seo}, \bibinfo{person}{Joice Tang}, \bibinfo{person}{Ido Roll}, \bibinfo{person}{Sidney Fels}, {and} \bibinfo{person}{Dongwook Yoon}.} \bibinfo{year}{2021}\natexlab{}.
\newblock \showarticletitle{The impact of artificial intelligence on learner–instructor interaction in online learning}.
\newblock \bibinfo{journal}{\emph{International Journal of Educational Technology in Higher Education}}  \bibinfo{volume}{18} (\bibinfo{year}{2021}), \bibinfo{pages}{1--23}.
\newblock


\bibitem[UNESCO(2023)]%
        {UNESCO2023}
\bibfield{author}{\bibinfo{person}{UNESCO}.} \bibinfo{year}{2023}\natexlab{}.
\newblock \bibinfo{title}{AI and education: guidance for policy-makers}.
\newblock \bibinfo{howpublished}{\url{https://unesdoc.unesco.org/ark:/48223/pf0000388367}}.
\newblock
\newblock
\shownote{Accessed: [Insert Date]}.


\bibitem[Wang et~al\mbox{.}(2023)]%
        {Wang2023}
\bibfield{author}{\bibinfo{person}{Yuntao Wang}, \bibinfo{person}{Yanghe Pan}, \bibinfo{person}{Miao Yan}, \bibinfo{person}{Zhou Su}, {and} \bibinfo{person}{Tom~H. Luan}.} \bibinfo{year}{2023}\natexlab{}.
\newblock \showarticletitle{A survey on ChatGPT: AI–generated contents, challenges, and solutions}.
\newblock \bibinfo{journal}{\emph{IEEE Open Journal of the Computer Society}}  \bibinfo{volume}{4} (\bibinfo{year}{2023}), \bibinfo{pages}{280--302}.
\newblock


\bibitem[West et~al\mbox{.}(2018)]%
        {West2018}
\bibfield{author}{\bibinfo{person}{Adam West}, \bibinfo{person}{John Clifford}, {and} \bibinfo{person}{David Atkinson}.} \bibinfo{year}{2018}\natexlab{}.
\newblock \showarticletitle{"Alexa, build me a brand" An Investigation into the impact of Artificial Intelligence on Branding}.
\newblock \bibinfo{journal}{\emph{The Business \& Management Review}} \bibinfo{volume}{9}, \bibinfo{number}{3} (\bibinfo{year}{2018}), \bibinfo{pages}{321--330}.
\newblock


\bibitem[Wu et~al\mbox{.}(2021)]%
        {Wu2021}
\bibfield{author}{\bibinfo{person}{Zhuohao Wu}, \bibinfo{person}{Danwen Ji}, \bibinfo{person}{Kaiwen Yu}, \bibinfo{person}{Xianxu Zeng}, \bibinfo{person}{Dingming Wu}, {and} \bibinfo{person}{Mohammad Shidujaman}.} \bibinfo{year}{2021}\natexlab{}.
\newblock \showarticletitle{AI creativity and the human-AI co-creation model}. In \bibinfo{booktitle}{\emph{Human-Computer Interaction. Theory, Methods and Tools: Thematic Area, HCI 2021, Held as Part of the 23rd HCI International Conference, HCII 2021, Virtual Event, July 24–29, 2021, Proceedings, Part I}} \emph{(\bibinfo{series}{Lecture Notes in Computer Science}, Vol.~\bibinfo{volume}{23})}. \bibinfo{publisher}{Springer International Publishing}, \bibinfo{pages}{171--190}.
\newblock


\bibitem[Yin et~al\mbox{.}(2023)]%
        {Yin2023}
\bibfield{author}{\bibinfo{person}{Hu Yin}, \bibinfo{person}{Zipeng Zhang}, {and} \bibinfo{person}{Yuanyuan Liu}.} \bibinfo{year}{2023}\natexlab{}.
\newblock \showarticletitle{The exploration of integrating the Midjourney artificial intelligence generated content tool into design systems to direct designers towards future-oriented innovation}.
\newblock \bibinfo{journal}{\emph{Systems}} \bibinfo{volume}{11}, \bibinfo{number}{12} (\bibinfo{year}{2023}), \bibinfo{pages}{566}.
\newblock


\bibitem[Zamfirescu-Pereira et~al\mbox{.}(2023)]%
        {ZamfirescuPereira2023}
\bibfield{author}{\bibinfo{person}{J.~Diego Zamfirescu-Pereira}, \bibinfo{person}{Richmond~Y. Wong}, \bibinfo{person}{Bj\"{o}rn Hartmann}, {and} \bibinfo{person}{Qian Yang}.} \bibinfo{year}{2023}\natexlab{}.
\newblock \showarticletitle{Why Johnny can't prompt: how non-AI experts try (and fail) to design LLM prompts}. In \bibinfo{booktitle}{\emph{Proceedings of the 2023 CHI Conference on Human Factors in Computing Systems}}. \bibinfo{pages}{1--21}.
\newblock


\bibitem[Zhou et~al\mbox{.}(2023)]%
        {Zhou2023}
\bibfield{author}{\bibinfo{person}{Jie Zhou}, \bibinfo{person}{Pei Ke}, \bibinfo{person}{Xipeng Qiu}, \bibinfo{person}{Minlie Huang}, {and} \bibinfo{person}{Junping Zhang}.} \bibinfo{year}{2023}\natexlab{}.
\newblock \showarticletitle{ChatGPT: potential, prospects, and limitations}.
\newblock \bibinfo{journal}{\emph{Frontiers of Information Technology \& Electronic Engineering}} (\bibinfo{year}{2023}), \bibinfo{pages}{1--6}.
\newblock


\bibitem[Šedlbauer et~al\mbox{.}(2024)]%
        {Sedlbauer2024}
\bibfield{author}{\bibinfo{person}{Josef Šedlbauer}, \bibinfo{person}{Jan Činčera}, \bibinfo{person}{Martin Slavík}, {and} \bibinfo{person}{Adéla Hartlová}.} \bibinfo{year}{2024}\natexlab{}.
\newblock \showarticletitle{Students' reflections on their experience with ChatGPT}.
\newblock \bibinfo{journal}{\emph{Journal of Computer Assisted Learning}} \bibinfo{volume}{40}, \bibinfo{number}{4} (\bibinfo{year}{2024}), \bibinfo{pages}{1526--1534}.
\newblock


\end{thebibliography}

\appendix
\onecolumn
\section{Student Prompt Histories and Observed Behaviors}

\newlength{\promptspacing}
\newlength{\continuationspacing}
\newlength{\sectspacing}
\setlength{\promptspacing}{1ex}
\setlength{\continuationspacing}{0.3ex}
\setlength{\sectspacing}{1ex}

\newlength{\numberwidth}
\newlength{\promptwidth}
\newlength{\behaviorwidth}
\setlength{\numberwidth}{0.02\textwidth}  
\setlength{\promptwidth}{0.55\textwidth}  
\setlength{\behaviorwidth}{0.32\textwidth}

\begin{table}[H]

\label{tab:student-prompts}
\begin{tabular}{p{\numberwidth}p{\promptwidth}p{\behaviorwidth}}
\toprule
& \textbf{Prompts} & \textbf{Observed Behavior} \\
\midrule
& \multicolumn{2}{l}{\textbf{S1 begins with specific materials}} \\[\promptspacing]
P1 & Which items of clothing can be made with 100\% hemp fabric? And with 100\% Tencel? 
    & Demonstrates initial domain knowledge \\[\promptspacing]
P2 & What differences are there in the production of cotton jeans and hemp jeans from a sustainability perspective? 
    & Narrows down; shows curiosity \\[\promptspacing]
P3 & What are the most eco-friendly fibers for clothing? 
    & Returns to broader perspective \\[\promptspacing]
P4 & List the advantages of organic cotton, hemp and tencel 
    & Focuses on specific materials \\[\promptspacing]
P5 & Are all of these fabrics biodegradable? 
    & Confirmatory question \\[\promptspacing]
P6 & What are the basic production processes of creating hemp, cotton and tencel fabrics? And what are the basic steps to turn them into clothing? 
    & Shifts focus to production processes \\[\promptspacing]
P7 & Explain how ramie is a sustainable fabric, as well as its advantages and disadvantages 
    & Introduces new material (tangential) \\[\promptspacing]
P8 & How do you produce ramie fabric? 
    & Follows up on new material \\[\sectspacing]
\midrule
& \multicolumn{2}{l}{\textbf{S2 begins with design-related prompts}} \\[\promptspacing]
P1 & I want to make a tote bag designed for students, but can also be used for casual outings, with water bottle pockets and ... 
    & Demonstrates initial domain knowledge; design goals \\[\promptspacing]
P2 & What materials are commonly used for tote bags 
    & Returns to broader perspective \\[\promptspacing]
P3 & Which material choice would be most durable, water proof and more suitable for student 
    & Compare options \\[\promptspacing]
P4 & The internet shows that nylon and polyster blend fabrics are recommended; what do you think 
    & External source validation \\[\promptspacing]
P5 & What material can be used for the water bottle pockets, since it has to be flexible and stretchy for various sizes of water bottles 
    & Focus on specifics \\[\promptspacing]
P6 & What are the basic production steps for a bag 
    & Shifts focus to production process \\[\sectspacing]
\midrule
& \multicolumn{2}{l}{\textbf{S3 adopts a more broad to specific approach}} \\[\promptspacing]
P1 & What materials are mainly used for black/textured hoodies? 
    & Starts with broad question \\[\promptspacing]
P2 & What is the raw material used for graphic designs on hoodies? 
    & Shifts focus to specific component \\[\promptspacing]
P3 & How are plasticol ink and heat transfer vinyl used together? 
    & Shifts focus to specific materials \\[\promptspacing]
P4 & Can you list the steps of manufacturing hoodies with graphic designs? 
    & Shifts focus to production process \\
\bottomrule
\end{tabular}

  \vspace{2mm} 
  \parbox{160mm}{ 
      \footnotesize\textsuperscript{1} Students were asked to identify the raw materials and the production process of their brand merchandise to ensure that they are aligned with their brand designs. Hence all students shifted to production process at one point in their prompt histories.
  }
  \vspace{1mm} 
  \parbox{160mm}{ 
      \footnotesize\textsuperscript{2} Students with prior domain knowledge (S1 and S2) exhibited non-linear learning patterns, exploring both specific and broad concepts iteratively. All three students showed accelerated acquisition of domain-specific knowledge.
  }

\end{table}
\vspace{\baselineskip}
\vspace{\baselineskip}
\vspace{\baselineskip}
\vspace{\baselineskip}
\vspace{\baselineskip}
\vspace{\baselineskip}
\vspace{\baselineskip}
\vspace{\baselineskip}
\vspace{\baselineskip}
\vspace{\baselineskip}
\vspace{\baselineskip}
\vspace{\baselineskip}

\twocolumn
\section{Mini-Presentation 4 Assignment Instructions}

Overview: The objective of this assignment is to conduct a comprehensive research project on one of the suppliers or factories listed on 1688.com. For those not proficient in Chinese, the website can be translated into English when accessed via a computer. As an alternative, you may use Alibaba.com, which offers language selection at the top right corner of the homepage. Please note that suppliers on Alibaba.com may have a limited product range and higher prices due to their international shipping capabilities.

Your Presentation should include:

\begin{enumerate}
    \item \textbf{A very short introduction into a product category you are interested in: }this can be a popular consumer good in your home country; or a product you would like to introduce for your own brand or for a certain influencer (for example, a custom-designed water bottle for the gym goers); or simply something you are curious about. (suggested duration 1min)
    
    \item \textbf{2.	Basic research into the raw materials: } list what types of raw materials you need for the product. Make it as specific as possible. For example, instead of just mentioning “plastic”, you should identify the exact type such as PVC (Polyvinyl Chloride) or Nylon. If you want to maintain high safety standards for the product, you need to be careful with what materials you use. Use pictures as well to illustrate the materials. (suggested duration 1min)
    
    \item \textbf{Basic research into the production process: } what types of special production processes are involved in creating the product. Detail the methods used, such as sewing buttons onto jeans either by hand or machine, stringing beads into a necklace by hand. Use images to illustrate the production process or craftsman ship. (suggested duration 1min)

    \item \textbf{Find a few suppliers/factories on 1688.com (or alibaba.com) who can provide such materials and production service.} Sometimes, you source different materials from different suppliers and get them assembled by another factory (if cheaper). You can also choose to buy kits with higher prices from a particular supplier. Tell us why you believe such factories are good choices, for example, low prices, authenticity (evidenced by factory photos), many followers, contracts with other big-name brands, long years of service, experience with international e-commerce. (suggested 2min).

\end{enumerate}

\noindent Tips:

\begin{itemize}
    \item Your initial research for Questions 2 and 3 DOES NOT need to be 100\% accurate. The primary goal is for you to engage with new and unfamiliar topics, so the emphasis will be on your research effort rather than precision. This approach will be reflected in the grading rubric.
    \item I highly recommend you use ChatGPT to you get started with Question 2 and 3. Remember, it is very rare to get the perfect answers with the first attempt. Start with broad questions to establish a foundation, such as, "What types of plastics are used for making packaging?" From the general answers you receive, refine your questions to dig deeper. For instance, you could follow up with, "What are the different types of PVC, and what are their common applications, advantages, and disadvantages?" You should practice your prompting skills through this exercise.
    \item Your presentation time should not exceed 5min. I will put a timer by the desk in the next class. You cannot continue beyond the 5min cutoff.
\end{itemize}

\noindent Grading Rubrics (weighed equally)
\begin{itemize}
    \item Detail and Specificity - The research provides specific details about the raw materials, the production processes, avoiding generic terms.
    \item Visual Illustration with engaging pictures
    \item Safety (or ESG) considerations
    \item Research depth - Evidence of thorough research in finding suitable suppliers or factories with clear justification for choices.
    \item Presentation Organization, Delivery and Time Management – well structured, logically organized, confident delivery within the time limit.
\end{itemize}

\end{document}